\documentclass{article}

\usepackage{PRIMEarxiv}
\usepackage{tikz}
\usepackage{pgfplots}

\usepackage[utf8]{inputenc} 
\usepackage[T1]{fontenc}    
\usepackage{hyperref}       
\usepackage{url}            
\usepackage{booktabs}       
\usepackage{amsfonts} 
\usepackage{amsmath}
\usepackage{nicefrac}       
\usepackage{microtype}      
\usepackage{lipsum}
\usepackage{fancyhdr}       
\usepackage{graphicx}  

\usepackage[]{biblatex}
\graphicspath{{media/}}     

\title{Cyclical Log Annealing as a Learning Rate scheduler}

\author{
  Philip Naveen \\
  University of Virginia \\
  Charlottesville\\
  \texttt{philipnaveen@email.virginia.edu}
}

\begin{document}

\maketitle

\begin{abstract}
A learning rate scheduler is a predefined set of instructions for varying search stepsizes during model training processes. This paper introduces a new logarithmic method using harsh restarting of step sizes through stochastic gradient descent. Cyclical log annealing implements the restart pattern more aggressively to maybe allow the usage of more greedy algorithms on the online convex optimization framework. The algorithm was tested on the CIFAR-10 image datasets, and seemed to perform analogously with cosine annealing on large transformer-enhanced residual neural networks. Future experiments would involve testing the scheduler in generative adversarial networks and finding the best parameters for the scheduler with more experiments.
\end{abstract}


\section{Introduction}

\subsection{Gradient Based Optimization}

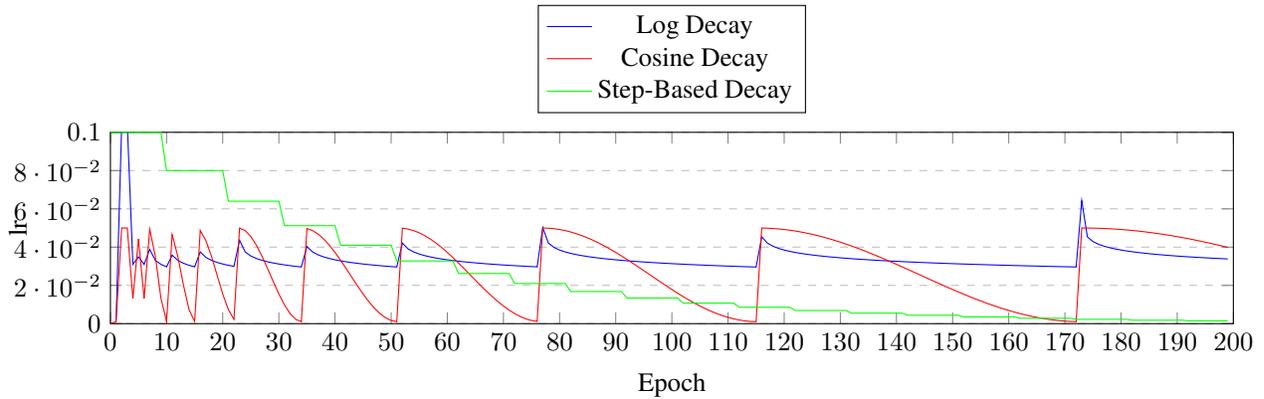
\begin{figure*}[h]
\centering
\begin{tikzpicture}
\begin{axis}[
    xlabel={Epoch},
    ylabel={lr},
    xmin=0, xmax=200,
    ymin=0, ymax=0.1,
    xtick={0, 10, 20, 30, 40, 50, 60, 70, 80, 90, 100, 110, 120, 130, 140, 150, 160, 170, 180, 190, 200},
    legend style={at={(.5 , 1.1)},anchor=south},
    ymajorgrids=true,
    grid style=dashed,
    cycle list name=color list,
    width=1\textwidth,
    height=.25\textwidth
]

\addplot[blue] coordinates{

(1,0.001)
(2,0.10306005729366875)
(3,0.10449813590929016)
(4,0.03099813590574342)
(5,0.034894626149073925)
(6,0.030998135906925666)
(7,0.038791116388857697)
(8,0.033082861364796434)
(9,0.030998135907713827)
(10,0.029693912358114173)
(11,0.03578597265241871)
(12,0.03308286136574223)
(13,0.03156682932121123)
(14,0.030508152816026708)
(15,0.02969391235847794)
(16,0.03748689284859475)
(17,0.03464117991200084)
(18,0.033082861366372764)
(19,0.03200390324328091)
(20,0.031177755973820213)
(21,0.030508152816331804)
(22,0.029945113829763413)
(23,0.04357895313401295)
(24,0.03748689285004982)
(25,0.03536197729420539)
(26,0.0340424164741489)
(27,0.033082861366793115)
(28,0.032328402628624825)
(29,0.03170662035242677)
(30,0.03117775597406587)
(31,0.030717615197245144)
(32,0.030310374012803315)
(33,0.02994511382993696)
(34,0.029613982479970626)
(35,0.040295491117864325)
(36,0.037486892851019867)
(37,0.03593936902786785)
(38,0.034865554367316774)
(39,0.0340424164745162)
(40,0.03337478859684079)
(41,0.03281314565477574)
(42,0.032328402628851366)
(43,0.031902015066293116)
(44,0.03152143045238971)
(45,0.031177755974229646)
(46,0.03086446104100317)
(47,0.030576608273801705)
(48,0.03031037401293156)
(49,0.03006273665549154)
(50,0.02983126691988725)
(51,0.029613982480076007)
(52,0.04212215818945984)
(53,0.03909574458716842)
(54,0.037486892851666564)
(55,0.0363841609435975)
(56,0.03554419332623501)
(57,0.034865554367625604)
(58,0.03429614613856393)
(59,0.033805632792009675)
(60,0.03337478859704363)
(61,0.03299065826699204)
(62,0.03264409560207606)
(63,0.03232840262900239)
(64,0.032038526955323596)
(65,0.0317705632816035)
(66,0.031521430452510005)
(67,0.031288654724975676)
(68,0.031070219907044082)
(69,0.030864444104100317)
(70,0.030669987278837674)
(71,0.03048562482823857)
(72,0.03031037401301705)
(73,0.030143376452206383)
(74,0.029983889626191286)
(75,0.02983126691996194)
(76,0.0296849417864169)
(77,0.05022714087015749)
(78,0.042122158191052764)
(79,0.039848946884259474)
(80,0.038474814925100624)
(81,0.037486892852097706)
(82,0.03671505077011663)
(83,0.036081524091687786)
(84,0.035544193326484314)
(85,0.03507765982883431)
(86,0.034665420850436605)
(87,0.034296146138739284)
(88,0.03396171933968506)
(89,0.03365612679307714)
(90,0.033374788597178864)
(91,0.033114136197917955)
(92,0.03287133492468326)
(93,0.03264409560218612)
(94,0.03243054298314576)
(95,0.03222912160096322)
(96,0.032038526955416376)
(97,0.03185765426754268)
(98,0.03168555968144393)
(99,0.0315214304525902)
(100,0.031364561735059054)
(101,0.031214338288929624)
(102,0.031070219907114695)
(103,0.030931729689455255)
(104,0.030798444521558027)
(105,0.030669987278900752)
(106,0.030546020394139235)
(107,0.03042624051120968)
(108,0.03031037401307405)
(109,0.03019817325718895)
(110,0.0300894133884163)
(111,0.029983889626243272)
(112,0.02988141494404873)
(113,0.02978181807433249)
(114,0.029684941786464678)
(115,0.029590641393463776)
(116,0.0453616461455418)
(117,0.04212215819211471)
(118,0.040457771927553066)
(119,0.03932947942606462)
(120,0.03847481492548037)
(121,0.037786630671267814)
(122,0.03721052401517693)
(123,0.03671505077034784)
(124,0.0362803861130691)
(125,0.035893222565778114)
(126,0.03554419332665052)
(127,0.03522645520231666)
(128,0.03493485629780856)
(129,0.034665420850566335)
(130,0.03441501640810521)
(131,0.03418113099463863)
(132,0.03396171933979144)
(133,0.03375509397673159)
(134,0.033559846361205764)
(135,0.03337478859726902)
(136,0.03319890962931354)
(137,0.03303134179404946)
(138,0.032871334924761485)
(139,0.03271823604966542)
(140,0.03257147329407431)
(141,0.032430542983214836)
(142,0.032294999211167054)
(143,0.03216444533084016)
(144,0.03203852695547823)
(145,0.03191692616054815)
(146,0.03179935664711477)
(147,0.03168555968149993)
(148,0.03157530066635021)
(149,0.03146836622883358)
(150,0.03136456173511021)
(151,0.03126370915831768)
(152,0.03116564524139961)
(153,0.031070219907161772)
(154,0.030977294876676782)
(155,0.030886742464109485)
(156,0.030798444521601628)
(157,0.030712291512338365)
(158,0.030628181693552248)
(159,0.030546020394179838)
(160,0.030465719374309273)
(161,0.03038719625555097)
(162,0.03031037401311205)
(163,0.030235180521723522)
(164,0.030161548148710606)
(165,0.030089413388452006)
(166,0.030018716533277107)
(167,0.02994940137652742)
(168,0.029881414944082398)
(169,0.029814707251136656)
(170,0.029749231081431265)
(171,0.029684941786496537)
(172,0.02962179710276828)
(173,0.06425478261627984)
(174,0.045361646147306554)
(175,0.04291186195643371)
(176,0.04147666382419469)
(177,0.040457771927995864)
(178,0.0396672051079103)
(179,0.039021135017607324)
(180,0.038474814925733535)
(181,0.03800152307441212)
(182,0.03758401786448599)
(183,0.03721052401535418)
(184,0.03687264111680635)
(185,0.036564165677857995)
(186,0.036280386113205464)
(187,0.03601763965207199)
(188,0.035773022433654975)
(189,0.03554419332676132)
(190,0.03532923731101927)
(191,0.035126567956437565)
(192,0.03493485629790188)
(193,0.03475297797019048)
(194,0.03457997325144101)
(195,0.0344150164081858)
(196,0.03425739185874535)
(197,0.03410647541235707)
(198,0.03396171933986236)
(199,0.033822640373657835)

};

      \addplot[red, ] coordinates {
        (0,0.0001)
(1,0.001)
(2,0.05)
(3,0.05)
(4,0.01325)
(5,0.04426808885641496)
(6,0.01325)
(7,0.04933959932920569)
(8,0.035203954267959345)
(9,0.01325)
(10,0.001165660235322396)
(11,0.046476089038491535)
(12,0.035203954267959345)
(13,0.020313255130379446)
(14,0.007356671678829392)
(15,0.001165660235322396)
(16,0.04862887534955585)
(17,0.043428962921731)
(18,0.035203954267959345)
(19,0.025341628633784793)
(20,0.0155060244906595)
(21,0.007356671678829392)
(22,0.002268584277056129)
(23,0.049955423587973745)
(24,0.04862887534955585)
(25,0.04555407365462912)
(26,0.04096343502193874)
(27,0.035203954267959345)
(28,0.028710976042340484)
(29,0.021975288221899206)
(30,0.0155060244906595)
(31,0.00979218020651387)
(32,0.005265650473255257)
(33,0.0022685842770561264)
(34,0.001027522306026121)
(35,0.04971652763301341)
(36,0.04862887534955585)
(37,0.046761477831061604)
(38,0.044177290767980756)
(39,0.04096343502193874)
(40,0.03722825951462751)
(41,0.03309768845376722)
(42,0.028710976042340484)
(43,0.02421601179285453)
(44,0.019764334718868196)
(45,0.01550602449065951)
(46,0.011584641789825434)
(47,0.008132388438958683)
(48,0.005265650473255266)
(49,0.0030810744107612515)
(50,0.001652309002207521)
(51,0.001027522306026121)
(52,0.04989867808500851)
(53,0.0494434370607239)
(54,0.04862887534955585)
(55,0.04746721713104985)
(56,0.045975895484171886)
(57,0.044177290767980756)
(58,0.04209839475815237)
(59,0.03977040557968092)
(60,0.03722825951462751)
(61,0.03451010671110728)
(62,0.031656738661583854)
(63,0.028710976042340484)
(64,0.025717026100860386)
(65,0.02271981923484455)
(66,0.019764334718868196)
(67,0.016894925697538785)
(68,0.014154653575026436)
(69,0.011584641789825434)
(70,0.009223458672690255)
(71,0.007106538649236358)
(72,0.005265650473255257)
(73,0.003728420471000436)
(74,0.0025179179511474487)
(75,0.001652309002207521)
(76,0.0011445838728787244)
(77,0.04999895020104838)
(78,0.04989867808500851)
(79,0.04963555825172601)
(80,0.049211346880943974)
(81,0.04862887534955585)
(82,0.04789203133372649)
(83,0.047005732860798365)
(84,0.045975895484171886)
(85,0.04480939280024973)
(86,0.04351401057097245)
(87,0.04209839475815238)
(88,0.04057199381644789)
(89,0.03894499563014055)
(90,0.03722825951462751)
(91,0.03543324373648143)
(92,0.033571929035841906)
(93,0.03165673866158385)
(94,0.029700455452984324)
(95,0.02771613652132165)
(96,0.025717026100860386)
(97,0.023716467150893365)
(98,0.02172781229884888)
(99,0.019764334718868196)
(100,0.017839139540690007)
(101,0.015965076380138606)
(102,0.014154653575026425)
(103,0.012419954698899123)
(104,0.01077255790984773)
(105,0.009223458672690255)
(106,0.007782996370309436)
(107,0.006460785293975469)
(108,0.005265650473255266)
(109,0.004205568773808233)
(110,0.003287615656208273)
(111,0.0025179179511474487)
(112,0.0019016129662208416)
(113,0.0014428141972322255)
(114,0.0011445838728787244)
(115,0.0010089125160634292)
(116,0.049983684110258185)
(117,0.04989867808500851)
(118,0.0497412729142485)
(119,0.049511935672457444)
(120,0.049211346880943974)
(121,0.04884039848850876)
(122,0.04840019122473013)
(123,0.04789203133372649)
(124,0.04731742669808735)
(125,0.0466780823644747)
(126,0.045975895484171886)
(127,0.045212949683592824)
(128,0.04439150888145638)
(129,0.04351401057097245)
(130,0.04258305858697355)
(131,0.04160141537945465)
(132,0.04057199381644789)
(133,0.0394978485405561)
(134,0.03838216690479316)
(135,0.03722825951462751)
(136,0.03603955040429389)
(137,0.034819566876523586)
(138,0.033571929035841906)
(139,0.03230033904649139)
(140,0.031008570146856023)
(141,0.029700455452984324)
(142,0.028379876584435172)
(143,0.027050752146197295)
(144,0.02571702610086038)
(145,0.02438265606554168)
(146,0.023051601568294916)
(147,0.021727812298848886)
(148,0.020415216388539624)
(149,0.019117708754213685)
(150,0.017839139540690007)
(151,0.016583302696075643)
(152,0.015353924713836089)
(153,0.014154653575026436)
(154,0.01298904792349543)
(155,0.011860566506183334)
(156,0.010772557909847724)
(157,0.009728250624671888)
(158,0.008730743464240447)
(159,0.007782996370309436)
(160,0.0068878216296560995)
(161,0.006047875529071042)
(162,0.005265650473255266)
(163,0.004543467589010813)
(164,0.0038834698376710676)
(165,0.003287615656208276)
(166,0.002757673145887443)
(167,0.0022952148257107566)
(168,0.0019016129662208416)
(169,0.0015780355175090432)
(170,0.0013254426435117736)
(171,0.0011445838728787244)
(172,0.0010359958748673956)
(173,0.04999999961472731)
(174,0.049983684110258185)
(175,0.049935074662980763)
(176,0.04985423538868348)
(177,0.0497412729142485)
(178,0.049596336237011054)
(179,0.04941961652823215)
(180,0.049211346880943974)
(181,0.04897180200250054)
(182,0.048701297852239075)
(183,0.04840019122473013)
(184,0.04806887927916612)
(185,0.04770779901550887)
(186,0.04731742669808735)
(187,0.04689827722740574)
(188,0.04645090346099048)
(189,0.045975895484171886)
(190,0.04547387983176262)
(191,0.04494551866165913)
(192,0.04439150888145638)
(193,0.04381258122922782)
(194,0.043209499309682875)
(195,0.04258305858697355)
(196,0.04193408533547836)
(197,0.04126343554994767)
(198,0.04057199381644789)
(199,0.039860672145593766)
};

\addplot[
    color=green
    ]
    coordinates {
(0,0.1)
(1,0.1)
(2,0.1)
(3,0.1)
(4,0.1)
(5,0.1)
(6,0.1)
(7,0.1)
(8,0.1)
(9,0.1)
(10,0.08000000000000002)
(11,0.08000000000000002)
(12,0.08000000000000002)
(13,0.08000000000000002)
(14,0.08000000000000002)
(15,0.08000000000000002)
(16,0.08000000000000002)
(17,0.08000000000000002)
(18,0.08000000000000002)
(19,0.08000000000000002)
(20,0.08000000000000002)
(21,0.06400000000000002)
(22,0.06400000000000002)
(23,0.06400000000000002)
(24,0.06400000000000002)
(25,0.06400000000000002)
(26,0.06400000000000002)
(27,0.06400000000000002)
(28,0.06400000000000002)
(29,0.06400000000000002)
(30,0.06400000000000002)
(31,0.051200000000000016)
(32,0.051200000000000016)
(33,0.051200000000000016)
(34,0.051200000000000016)
(35,0.051200000000000016)
(36,0.051200000000000016)
(37,0.051200000000000016)
(38,0.051200000000000016)
(39,0.051200000000000016)
(40,0.051200000000000016)
(41,0.04096000000000002)
(42,0.04096000000000002)
(43,0.04096000000000002)
(44,0.04096000000000002)
(45,0.04096000000000002)
(46,0.04096000000000002)
(47,0.04096000000000002)
(48,0.04096000000000002)
(49,0.04096000000000002)
(50,0.04096000000000002)
(51,0.03276800000000016)
(52,0.032768000000000016)
(53,0.032768000000000016)
(54,0.032768000000000016)
(55,0.032768000000000016)
(56,0.032768000000000016)
(57,0.032768000000000016)
(58,0.032768000000000016)
(59,0.032768000000000016)
(60,0.032768000000000016)
(61,0.032768000000000016)
(62,0.026214400000000013)
(63,0.026214400000000013)
(64,0.026214400000000013)
(65,0.026214400000000013)
(66,0.026214400000000013)
(67,0.026214400000000013)
(68,0.026214400000000013)
(69,0.026214400000000013)
(70,0.026214400000000013)
(71,0.026214400000000013)
(72,0.02097152000000001)
(73,0.02097152000000001)
(74,0.02097152000000001)
(75,0.02097152000000001)
(76,0.02097152000000001)
(77,0.02097152000000001)
(78,0.02097152000000001)
(79,0.02097152000000001)
(80,0.02097152000000001)
(81,0.02097152000000001)
(82,0.016777216000000008)
(83,0.016777216000000008)
(84,0.016777216000000008)
(85,0.016777216000000008)
(86,0.016777216000000008)
(87,0.016777216000000008)
(88,0.016777216000000008)
(89,0.016777216000000008)
(90,0.016777216000000008)
(91,0.016777216000000008)
(92,0.013421772800000007)
(93,0.013421772800000007)
(94,0.013421772800000007)
(95,0.013421772800000007)
(96,0.013421772800000007)
(97,0.013421772800000007)
(98,0.013421772800000007)
(99,0.013421772800000007)
(100,0.013421772800000007)
(101,0.013421772800000007)
(102,0.010737418240000006)
(103,0.010737418240000006)
(104,0.010737418240000006)
(105,0.010737418240000006)
(106,0.010737418240000006)
(107,0.010737418240000006)
(108,0.010737418240000006)
(109,0.010737418240000006)
(110,0.010737418240000006)
(111,0.010737418240000006)
(112,0.008589934592000005)
(113,0.008589934592000005)
(114,0.008589934592000005)
(115,0.008589934592000005)
(116,0.008589934592000005)
(117,0.008589934592000005)
(118,0.008589934592000005)
(119,0.008589934592000005)
(120,0.008589934592000005)
(121,0.008589934592000005)
(122,0.0068719476736000045)
(123,0.0068719476736000045)
(124,0.0068719476736000045)
(125,0.0068719476736000045)
(126,0.0068719476736000045)
(127,0.0068719476736000045)
(128,0.0068719476736000045)
(129,0.0068719476736000045)
(130,0.0068719476736000045)
(131,0.0068719476736000045)
(132,0.005497558138880004)
(133,0.005497558138880004)
(134,0.005497558138880004)
(135,0.005497558138880004)
(136,0.005497558138880004)
(137,0.005497558138880004)
(138,0.005497558138880004)
(139,0.005497558138880004)
(140,0.005497558138880004)
(141,0.005497558138880004)
(142,0.004398046511104004)
(143,0.004398046511104004)
(144,0.004398046511104004)
(145,0.004398046511104004)
(146,0.004398046511104004)
(147,0.004398046511104004)
(148,0.004398046511104004)
(149,0.004398046511104004)
(150,0.004398046511104004)
(151,0.004398046511104004)
(152,0.0035184372088832034)
(153,0.0035184372088832034)
(154,0.0035184372088832034)
(155,0.0035184372088832034)
(156,0.0035184372088832034)
(157,0.0035184372088832034)
(158,0.0035184372088832034)
(159,0.0035184372088832034)
(160,0.0035184372088832034)
(161,0.0035184372088832034)
(162,0.002814749767106563)
(163,0.002814749767106563)
(164,0.002814749767106563)
(165,0.002814749767106563)
(166,0.002814749767106563)
(167,0.002814749767106563)
(168,0.002814749767106563)
(169,0.002814749767106563)
(170,0.002814749767106563)
(171,0.002814749767106563)
(172,0.0022517998136852503)
(173,0.0022517998136852503)
(174,0.0022517998136852503)
(175,0.0022517998136852503)
(176,0.0022517998136852503)
(177,0.0022517998136852503)
(178,0.0022517998136852503)
(179,0.0022517998136852503)
(180,0.0022517998136852503)
(181,0.0022517998136852503)
(182,0.0018014398509482003)
(183,0.0018014398509482003)
(184,0.0018014398509482003)
(185,0.0018014398509482003)
(186,0.0018014398509482003)
(187,0.0018014398509482003)
(188,0.0018014398509482003)
(189,0.0018014398509482003)
(190,0.0018014398509482003)
(191,0.0018014398509482003)
(192,0.0014411518807585604)
(193,0.0014411518807585604)
(194,0.0014411518807585604)
(195,0.0014411518807585604)
(196,0.0014411518807585604)
(197,0.0014411518807585604)
(198,0.0014411518807585604)
(199,0.0014411518807585604)
};

\legend{Log Decay, Cosine Decay, Step-Based Decay}

\end{axis}
\end{tikzpicture}

\caption{Learning Rate Schedulers; shared parameters are initial decay epochs=1, min decay lr=0.001, restart interval=1, restart interval multiplier=1.5, restart lr=0.05, warmup epochs=1, warmup start lr=0.0001 for annealing schedulers.}
\end{figure*}[h]

Deep neural networks (DNN) are becoming the staple for a growing majority of classification problems \cite{Alexnet, CIFAR-10first}, and are gaining use in generative algorithms such as the large language models. They typically perform well on training sets with thousands of parameters, and are able to produce difficult approximations easily. The main caveat to these algorithms is the long training time, which prompts the use of expensive computatioal equipment such as GPUs, TPUs \cite{SGDR}. Algorithms that speed up the training proccess are one of the most lucrative parts of the feild at the moment.

Suppose a model has $n$ parameters, is trained through the process of minimizing the function $f: \mathbb{R}^n \to \mathbb{R}$. Normally, we use iterative stochastic gradient descent at time step $t$ to adjust that parameter vector $\mathbf{x}_t \in \mathbb{R}^n$. We do this by taking gradient information $\nabla f_t (\mathbf{x}_t)$ which we get using some small batch $b$ of datapoints \cite{Adam, SGDR}.

There is stochastic gradient descent
\begin{align}
    \mathbf{x}_{t+1} = \mathbf{x}_{t} - \eta_t \nabla f_{t}(\mathbf{x}_{t}) 
\end{align}
and with second order information
\begin{equation}
    \mathbf{x}_{t+1} = \mathbf{x}_{t} - \eta_t \mathbf{H}_{t}^{-1} \nabla f_{t}(\mathbf{x}_{t}) 
\end{equation}
that is usually avoided using LBFGS because the difficulties behind stochasticity in $\nabla f_t (\mathbf{x}_t)$. Typically, the use of momentum based optimization and approximating second order gradient information is used to achieve a superior regret bound and avoid the difficulty and unpredictability that comes with conditioning a parameter space for a model \cite{Adam}. 

\subsection{Restarting Mechan}

In any form of optimizing multimodal functions regardless of the variation in divergence for each minimum or maximum, it is common there exists a single minima of interest. In his paper relating to global optimization, Guo et al suggested a multi armed bandit and ranking system. They further use variations of reward and restless algorithms to achieve better convergence results on objective functions with multiple extrema such as the Ackley, Griewank, and Rastrigin functions \cite{bandit}.

This paper takes a different approach more akin to Loshchilov and Hutter by using the idea of shifting the learning rate using a scheduling system to prevent the optimizer from getting stuck in saddle points and local minima \cite{SGDR, SchedulerLR}.

Restarts are usually used to avoid converging to local minima rather than dealing with multimodality. Suppose we have a collection of non-smooth, yet convex optimization problems. About a decade ago, Yang and Lin found that a linear convergence rate is achievable by geometrically decreasing the learning rate throughout training \cite{LinearConvergenceRates}.

Suppose we have $\eta_{min}^{i}$ and $\eta_{max}^{i}$ to define the range of learning rates, so that generally $\eta_{min}^{i}, \eta_{max}^{i} \in \mathbb{R}$. The $T_{cur}$ counts for the accumulation of epochs for training a model through a process such as backpropagation. In their paper on warm restarts to improve the convergence of stochastic gradient descent, Loshchilov and Hutter suggested cyclical cosine annealing
\begin{equation}
    \eta_t = \eta^{i}_{min} + \frac{1}{2} \left( \eta^{i}_{max} - \eta^{i}_{min}\right) \left( 1 + \cos(\frac{T_{cur}}{T_{i}} \pi) \right)
\end{equation}
for the purpose of rescheduling the learning rates in convolutional neural networks \cite{SGDR}.

This project introduces an alternative approach similar to theirs that uses logarithmic properties instead of those of cosine. We decide to vary the base of the logarithmic expansion using a difference in the current range of learning rates. Therefore, 
\begin{equation}
\eta_t = \left| \eta^{i}_{min} + \frac{1}{2} \left( \eta^{i}_{max} - \eta^{i}_{min}\right) \left( 1 + \log_{ \left( \underset{\eta_j \in \eta^{i}_{min}}{\max} \right)^{-1}}(\frac{T_{i}}{T_{cur}} \pi) \right) \right|
\end{equation}
is what we have introduced as a different restart mechanism. We introduced this alternative method with the reasoning that using the difference as a base in an asymptotic function such as a logarithm would help mitigate the range of $\eta$ without any edge cases such as division by 0.

A higher learning rate can give fast, yet unreliable convergence due to overfitting, but can be mitigated with dropout \cite{Alexnet}. Further, a high learning rate without proper batch normalization for prolonged periods during training can cause neurons using ReLU to die \cite{LeakyReLU}. Cosine annealing lowers $\eta_t$ at roughly the same pace as it increased it previously. This approach instead increases $\eta_t$ almost asymptotically for a select few $T$, and then anneals it at a logarithmic pace. The purpose of this was to rapidly diverge from a solution in a local minima in the span of a few $T$, and then reduce $\eta_t$ such that gradient descent can locate the global minima at a reasonable convergence rate such that there is no rapid overfitting from a prolonged high $\eta_t$.

\section{Experiments}

\subsection{CIFAR-10 Classification}

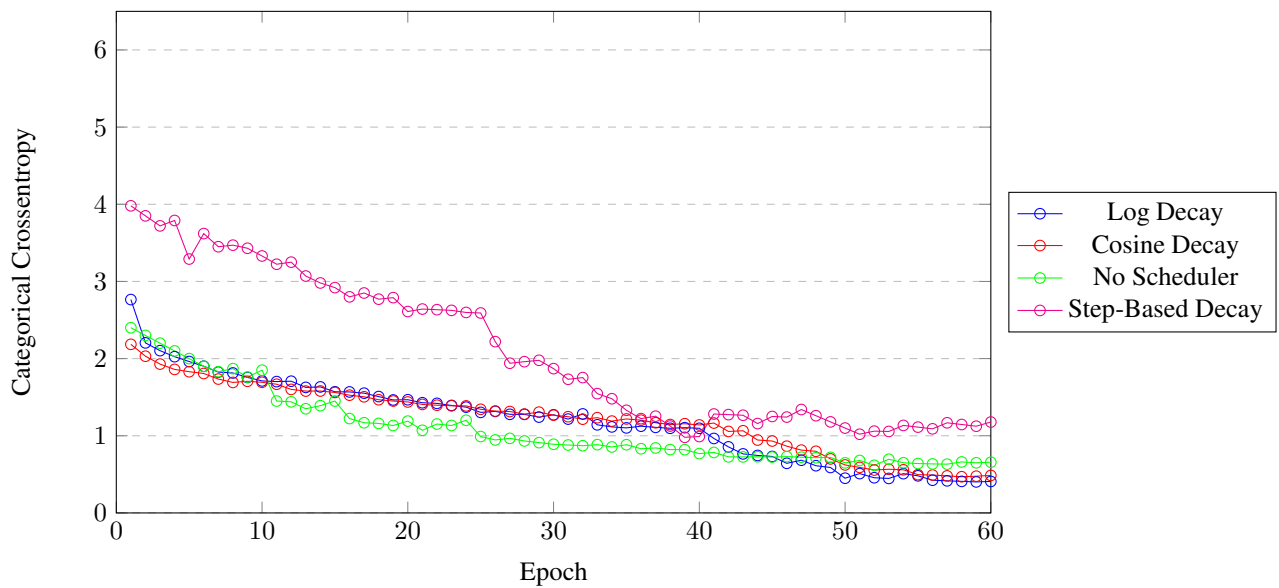
\begin{figure*}[h]
\centering
\begin{tikzpicture}
\begin{axis}[
    xlabel={Epoch},
    ylabel={Categorical Crossentropy},
    xmin=0, xmax=60,
    ymin=0, ymax=6.5,
    xtick={0, 10, 20, 30, 40, 50, 60},
    legend style={at={(1.02 ,0.5)},anchor=west},
    ymajorgrids=true,
    grid style=dashed,
    cycle list name=color list,
    width=0.8\textwidth,
    height=0.5\textwidth
]

\addplot[
    color=blue,
    mark=o,
    ]
    coordinates {
    (1,2.765)(2,2.204)(3,2.103)(4,2.025)(5,1.962)
    (6,1.901)(7,1.825)(8,1.814)(9,1.756)(10,1.711)
    (11,1.702)(12,1.707)(13,1.627)(14,1.633)(15,1.570)
    (16,1.568)(17,1.552)(18,1.508)(19,1.464)(20,1.467)
    (21,1.428)(22,1.420)(23,1.393)(24,1.371)(25,1.303)
    (26,1.316)(27,1.278)(28,1.277)(29,1.242)(30,1.272)
    (31,1.216)(32,1.283)(33,1.142)(34,1.115)(35,1.102)
    (36,1.124)(37,1.109)(38,1.097)(39,1.105)(40,1.097)
    (41,0.965)(42,0.856)(43,0.765)(44,0.745)(45,0.732)
    (46,0.643)(47,0.683)(48,0.612)(49,0.589)(50,0.450)
    (51,0.510)(52,0.456)(53,0.447)(54,0.509)(55,0.481)
    (56,0.424)(57,0.416)(58,0.409)(59,0.402)(60,0.408)
    };
    
\addplot[
    color=red,
    mark=o,
    ]
    coordinates {
    (1,2.185)(2,2.031)(3,1.929)(4,1.860)(5,1.830)
    (6,1.807)(7,1.735)(8,1.691)(9,1.706)(10,1.693)
    (11,1.668)(12,1.601)(13,1.579)(14,1.580)(15,1.561)
    (16,1.526)(17,1.505)(18,1.467)(19,1.450)(20,1.438)
    (21,1.406)(22,1.396)(23,1.389)(24,1.387)(25,1.343)
    (26,1.320)(27,1.314)(28,1.281)(29,1.306)(30,1.267)
    (31,1.247)(32,1.215)(33,1.234)(34,1.190)(35,1.218)
    (36,1.202)(37,1.169)(38,1.146)(39,1.156)(40,1.140)
    (41,1.165)(42,1.056)(43,1.065)(44,0.945)(45,0.932)
    (46,0.865)(47,0.814)(48,0.799)(49,0.700)(50,0.622)
    (51,0.590)(52,0.556)(53,0.566)(54,0.559)(55,0.496)
    (56,0.491)(57,0.480)(58,0.469)(59,0.476)(60,0.486)
    };

\addplot[
    color=green,
    mark=o,
    ]
    coordinates {
    (1,2.4)(2,2.3)(3,2.2)(4,2.1)(5,2.0)
    (6,1.9)(7,1.83)(8,1.87)(9,1.76)(10,1.85)
    (11,1.45)(12,1.44)(13,1.35)(14,1.39)(15,1.45)
    (16,1.223)(17,1.169)(18,1.16)(19,1.133)(20,1.19)
    (21,1.071)(22,1.152)(23,1.133)(24,1.201)(25,0.99)
    (26,0.946)(27,0.9665)(28,0.933)(29,0.911)(30,0.89)
    (31,0.88)(32,0.871)(33,0.885)(34,0.855)(35,0.884)
    (36,0.832)(37,0.842)(38,0.821)(39,0.820)(40,0.769)
    (41,0.784)(42,0.727)(43,0.726)(44,0.728)(45,0.724)
    (46,0.73)(47,0.722)(48,0.716)(49,0.720)(50,0.649)
    (51,0.68)(52,0.617)(53,0.696)(54,0.65)(55,0.64)
    (56,0.635)(57,0.632)(58,0.661)(59,0.650)(60,0.657)
    };

\addplot[
    color=magenta,
    mark=o,
    ]
    coordinates {
   (1,3.98)(2,3.850)(3,3.72)(4,3.79)(5,3.29)
    (6,3.62)(7,3.45)(8,3.47)(9,3.43)(10,3.330)
    (11,3.223)(12,3.25)(13,3.07)(14,2.98)(15,2.92)
    (16,2.80)(17,2.85)(18,2.77)(19,2.79)(20,2.61)
    (21,2.642)(22,2.634)(23,2.625)(24,2.598)(25,2.5910)
    (26,2.22)(27,1.94)(28,1.96)(29,1.978)(30,1.870)
    (31,1.732)(32,1.754)(33,1.546)(34,1.48)(35,1.330)
    (36,1.222)(37,1.254)(38,1.126)(39,0.98)(40,0.99)
    (41,1.282)(42,1.274)(43,1.266)(44,1.158)(45,1.250)
    (46,1.242)(47,1.34)(48,1.26)(49,1.18)(50,1.10)
    (51,1.02)(52,1.06)(53,1.056)(54,1.134)(55,1.112)
    (56,1.090)(57,1.168)(58,1.146)(59,1.124)(60,1.178)
    };

\legend{Log Decay, Cosine Decay, No Scheduler, Step-Based Decay}
    
\end{axis}
\end{tikzpicture}
\caption{The Effect of Learning Rate Schedulers on CIFAR-10 Image Classification using ResNet34}
\end{figure*}

To test the learning rate schedulers, we replicated a pretrained model on a benchmark dataset CIFAR-10 \cite{CIFAR-10first}. The model of 63.5 million parameters trained using stochastic gradient descent. The loss of interest was categorical crossentropy to ensure the model had some ambiguity in the output layer such that each output neuron is not strictly mututally exclusive. Each image was preprocessed using the pipeline created by Zagoruyko and Komodakis such that we augment each image using horizontal flips and randomized crops for 4x4 patches of pixels on the 32x32 image data. We also add pixel by mean, ZCA whitening, and global contrast normalization. We also use contrast stretching equalization \cite{Alexnet, SGDR}. 

SGD was used instead of Adam because it is less adaptable, and thus less compatable with a constant learning rate, as the gradient distribution can get distorted faster. The initial learning rate was set to $\eta_0$ = 0.0001, weight decay to 0.0005, dampening to 0, momentum to 0.9 and minibatch size to 128 \cite{SGDR}. We used three different schedulers and a control variable of no scheduler to compare. We used init decay epochs of 10, minimum decay for $\eta_t$ of 0.001, restart interval of 10, restart interval multiplier of 2.0, restart $\eta_t$ of 0.1, warmup epochs of 5, warmup start $\eta_t$ of 0.01. We trained ResNet34 \cite{ResNets} using the three schedulers for a total accumulation of 60 epochs due to constraints with GPU access.

In figure 3 the results of this is shown. These are the averages of each epoch, which is why the model's loss spikes for the restarts are not apparent, as the spikes sometimes stalled depending on the images in a particular set of batches in an epoch, for a set of epochs $T_n, T_{n+1} ... T_{n+10}$ between restarts we assigned. Thus, the highs and lows of the restarts cancelled out for each period for both cosine and log decay. The graph depicts the loss reduction of a large convolutional neural network when using learning rate schedulers through a budget of 60 epochs. When observing the graph, two things are immediately apparent:

\begin{itemize}
    \item [1] For dealing with categorical crossentropy loss, it seems that the loss is rather stable Poisson or deserialization loss when using learning rate schedulers, likely due to abstaining from using softmax, hardmax, or argmax in favor of a sigmoid dense layer for outputs.

    \item [2] Of the three schedulers and the control, it seems that log annealing and cosine annealing perform roughly the same for the majority of the training. However, we note that log annealing performs worse in the beginning and better at the end. This is because a high $\eta_t$ can result in higher variability in $\frac{\partial J}{\partial w^2_{j,i}}$ from inheritly more extreme predictions from less exposure to training data.
\end{itemize}

Considering these observations, there is a strong case that log annealing is at the very least capable of the same performance and exhibits similar properties to cosine annealing.

\subsection{Transformer CIFAR-10 Classification}

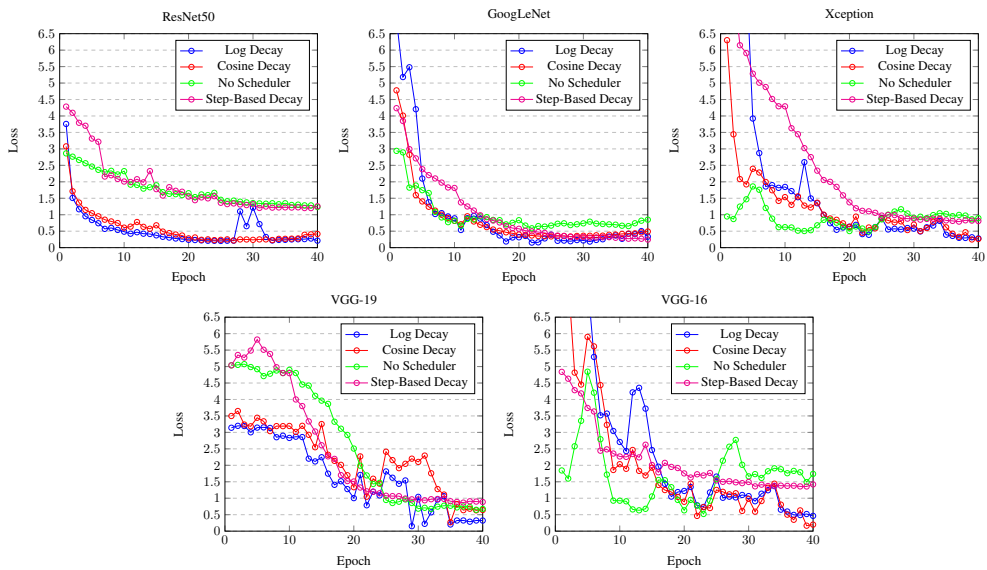
\begin{figure}[h]

    \centering

    \begin{tikzpicture}[scale=0.50]

        \begin{axis}[
            title = {ResNet50},
            xlabel={Epoch},
            ylabel={Loss},
            xmin=0, xmax=40,
            ymin=0, ymax=6.5,
            xtick={0, 10, 20, 30, 40, 50, 60},
            ytick={0, 0.5, 1, 1.5, 2, 2.5, 3, 3.5, 4, 4.5, 5, 5.5, 6, 6.5},
            legend pos=north east,
            ymajorgrids=true,
            grid style=dashed,
        ]
        
        \addplot[
            color=blue,
            mark=o,
            ]
            coordinates {(1, 3.7580558487521412)(2, 1.5096613187009416)(3, 1.1658924637395707)(4, 0.9536863602030917)(5, 0.8400313759322666)(6, 0.741469627260552)(7, 0.5706330016674593)(8, 0.5986808037285305)(9, 0.5387262947228558)(10, 0.4853666775557391)(11, 0.42405086978698325)(12, 0.4690172222416724)(13, 0.423992669292728)(14, 0.41505009936325993)(15, 0.37328067025565126)(16, 0.3293104242638249)(17, 0.3062060613785406)(18, 0.2782684996762239)(19, 0.2601157367191351)(20, 0.2374850066040483)(21, 0.2310099626231529)(22, 0.22017756409828773)(23, 0.20982794182570388)(24, 0.20797627606926977)(25, 0.2034340137496705)(26, 0.21035337288056494)(27, 0.21204597652530122)(28, 1.10098968331828293)(29, 0.65234607356398003)(30, 1.21553292927210746)(31, 0.7155279903970373)(32, 0.32356709998929897)(33, 0.21791911087310908)(34, 0.2343100297891194)(35, 0.23739289333734215)(36, 0.24971268830053947)(37, 0.2547274647647386)(38, 0.2594050548475264)(39, 0.2768326210968025)(40, 0.21266895259470893)
            };

            \addplot[
    color=red,
    mark=o,
    ]
    coordinates {
        (1, 3.076676367384268)(2, 1.712801140402587)(3, 1.3701726901729218)(4, 1.147204408437225)(5, 1.0462526379894735)(6, 0.9521338618506098)(7, 0.8492767118968885)(8, 0.7962202750663944)(9, 0.7473061492195254)(10, 0.6078138054890881)(11, 0.6381038764354214)(12, 0.7789260718103546)(13, 0.6326395946974778)(14, 0.5731391496921192)(15, 0.67848129229168156)(16, 0.4879592197370183)(17, 0.44071660153112814)(18, 0.3987813733909531)(19, 0.36814658335636656)(20, 0.28218266161051386)(21, 0.30931422818650966)(22, 0.23890658919332895)(23, 0.2520220927760421)(24, 0.22851683844931516)(25, 0.2374156415245412)(26, 0.25219653675128735)(27, 0.21926922184571705)(28, 0.2489352558732495)(29, 0.24846055603718324)(30, 0.2367641272428625)(31, 0.25010548010913956)(32, 0.25075129028806744)(33, 0.23154786124618194)(34, 0.26728568029334025)(35, 0.2680166006210808)(36, 0.2734780285895583)(37, 0.27984048530638607)(38, 0.3921137458858686)(39, 0.40315350410457696)(40, 0.41578264124819465)
    };

\addplot[
    color=green,
    mark=o,
    ]
    coordinates {
    (1,2.8639267392)(2,2.7639267392)(3,2.6639267392)(4,2.5639267392)(5,2.4639267392)
    (6,2.3639267392)(7,2.2939267392)(8,2.3339267392)(9,2.2239267392)(10,2.3239267392)
    (11,1.9139267392)(12,1.9039267392)(13,1.7939267392)(14,1.8339267392)(15,1.8939267392)
    (16,1.6869267392)(17,1.6329267392)(18,1.6239267392)(19,1.5969267392)(20,1.6539267392)
    (21,1.5349267392)(22,1.6159267392)(23,1.5969267392)(24,1.6649267392)(25,1.4539267392)
    (26,1.4099267392)(27,1.4294267392)(28,1.3969267392)(29,1.3749267392)(30,1.3539267392)
    (31,1.3439267392)(32,1.3349267392)(33,1.3489267392)(34,1.3189267392)(35,1.3479267392)
    (36,1.2959267392)(37,1.3059267392)(38,1.2849267392)(39,1.2839267392)(40,1.2329267392)
   
    };

\addplot[
    color=magenta,
    mark=o,
    ]
    coordinates {
    (1, 4.2844555861)(2, 4.0965964944)(3, 3.7927833289)(4, 3.7028828463)(5, 3.3129823638)
    (6, 3.2170497439)(7, 2.1603552286)(8, 2.1993074712)(9, 2.0877571236)(10, 2.00853054244)
    (11, 1.9946542999)(12, 2.07850794407)(13, 1.9840997006)(14, 2.3259839758)(15, 1.7774577251)
    (16, 1.5829241021)(17, 1.8366824726)(18, 1.7274765561)(19, 1.698493008)(20, 1.5513012189)
    (21, 1.4419997325)(22, 1.526646107)(23, 1.4972602434)(24, 1.5628254867)(25, 1.3630541546)
    (26, 1.3226709329)(27, 1.343139358)(28, 1.3112894677)(29, 1.2925026269)(30, 1.3201536281)
    (31, 1.20622612102)(32, 1.2491782526)(33, 1.2154475661)(34, 1.2215955875)(35, 1.223825893)
    (36, 1.2180401722)(37, 1.2245719827)(38, 1.2029550529)(39, 1.2020095497)(40, 1.2573165261)
    };

            \legend{Log Decay, Cosine Decay, No Scheduler, Step-Based Decay}
        \end{axis}
    \end{tikzpicture}
\begin{tikzpicture}[scale=0.50]

        \begin{axis}[
            title = {GoogLeNet},
            xlabel={Epoch},
            ylabel={Loss},
            xmin=0, xmax=40,
            ymin=0, ymax=6.5,
            xtick={0, 10, 20, 30, 40, 50, 60},
            ytick={0, 0.5, 1, 1.5, 2, 2.5, 3, 3.5, 4, 4.5, 5, 5.5, 6, 6.5},
            legend pos=north east,
            ymajorgrids=true,
            grid style=dashed,
        ]
        
\addplot[
    color=blue,
    mark=o,
    ]
    coordinates {
        (1, 6.940019269674102)(2, 5.181967004504285)(3, 5.4798602592432627)(4, 4.2054638969654012)(5, 2.0987621002108096)(6, 1.3793128904002203)(7, 1.0223572295546992)(8, 1.0480370786380138)(9, 0.9491733081924483)(10, 0.890089250918979)(11, 0.5368557254488505)(12, 0.8954688305879965)(13, 0.9644220579539825)(14, 0.8793061279479376)(15, 0.6855718274814776)(16, 0.4838750356722845)(17, 0.37013699857771297)(18, 0.18761064104680113)(19, 0.30863221427316554)(20, 0.3143834396402383)(21, 0.3562061133728632)(22, 0.15324407669340113)(23, 0.1610758772562395)(24, 0.28030821742258947)(25, 0.3924387075563181)(26, 0.207363701520768)(27, 0.21427282870725835)(28, 0.19505013406218735)(29, 0.23073465598184106)(30, 0.2178348851023221)(31, 0.18416097714716802)(32, 0.23382512403120728)(33, 0.25444930311188564)(34, 0.3368394326613725)(35, 0.3712609323808851)(36, 0.26816556674168795)(37, 0.4171556176350366)(38, 0.4080474769796591)(39, 0.5032434334950855)(40, 0.33414664046470244)
    };

\addplot[
    color=red,
    mark=o,
    ]
    coordinates {
    (1, 4.7783297989)(2, 4.0111860644)(3, 2.8227856619)(4, 1.595417569)(5, 1.4034549156)
    (6, 1.2421057831)(7, 1.1246499392)(8, 1.0022047273)(9, 0.9011352411)(10, 0.8134661709)
    (11, 0.7098369299)(12, 0.9528326728)(13, 0.7946144025)(14, 0.6943453343)(15, 0.6246553944)
    (16, 0.5504609315)(17, 0.5127235602)(18, 0.4659150487)(19, 0.4355346059)(20, 0.3981655656)
    (21, 0.3866083707)(22, 0.3693287314)(23, 0.3508860034)(24, 0.3478303122)(25, 0.3407306294)
    (26, 0.3525836467)(27, 0.3553341229)(28, 0.337074259)(29, 0.3555820122)(30, 0.3612939777)
    (31, 0.3612912206)(32, 0.3749892383)(33, 0.364549831)(34, 0.3923864435)(35, 0.3976868979)
    (36, 0.4178377094)(37, 0.426618366)(38, 0.4347074768)(39, 0.4636202437)(40, 0.4900810208)
    };

\addplot[
    color=green,
    mark=o,
    ]
    coordinates {
        (1, 2.939682224585221)(2, 2.893457059477313)(3, 1.8236936146965824)(4, 1.8857552130442921)(5, 1.7456280942517464)(6, 1.6593074682329114)(7, 1.08496452900579291)(8, 0.9173770042853331)(9, 0.7740959276693257)(10, 0.817099210372336)(11, 0.668578585455131)(12, 0.8714260297914637)(13, 0.8542166823666715)(14, 0.9506952054818864)(15, 0.9267365719611934)(16, 0.8421733278294515)(17, 0.8364986044763224)(18, 0.7074184874085692)(19, 0.7552606562885473)(20, 0.8338946835980655)(21, 0.6712564736935488)(22, 0.5882152689569266)(23, 0.6557414904178643)(24, 0.6392305182654609)(25, 0.670477307676666)(26, 0.7264712017786464)(27, 0.7392741913988261)(28, 0.6854939057543807)(29, 0.7107280811483178)(30, 0.7382870517595858)(31, 0.7890482360942087)(32, 0.7307780845284605)(33, 0.7182262638642137)(34, 0.7076735531552559)(35, 0.6976410303805011)(36, 0.6663788585429195)(37, 0.6550766260783639)(38, 0.7407777865769072)(39, 0.8189056049670488)(40, 0.8489470970388768)
    };

\addplot[
    color=magenta,
    mark=o,
    ]
    coordinates {
    (1, 4.2370342409)(2, 3.8478898903)(3, 2.9921876513)(4, 2.7117856563)(5, 2.3828847589)
    (6, 2.2032968336)(7, 2.1022223274)(8, 1.976139145)(9, 1.8282431165)(10, 1.813166473)
    (11, 1.3742028671)(12, 1.2510565331)(13, 1.1278313072)(14, 0.9827859489)(15, 0.8627988673)
    (16, 0.8233498535)(17, 0.7651082996)(18, 0.6570319867)(19, 0.5941711242)(20, 0.5732631046)
    (21, 0.5144384967)(22, 0.4944576428)(23, 0.4738672982)(24, 0.447154179)(25, 0.3935272865)
    (26, 0.3659169503)(27, 0.3530623731)(28, 0.3311067589)(29, 0.3240183693)(30, 0.3238717072)
    (31, 0.304416779)(32, 0.3094503937)(33, 0.3151817311)(34, 0.3185405253)(35, 0.2999642529)
    (36, 0.2931576446)(37, 0.2714089049)(38, 0.2838552718)(39, 0.2698610989)(40, 0.2442537851)
    };

            \legend{Log Decay, Cosine Decay, No Scheduler, Step-Based Decay}
        \end{axis}
    \end{tikzpicture}    
\begin{tikzpicture}[scale=0.50]

        \begin{axis}[
            title = {Xception},
            xlabel={Epoch},
            ylabel={Loss},
            xmin=0, xmax=40,
            ymin=0, ymax=6.5,
            xtick={0, 10, 20, 30, 40, 50, 60},
            ytick={0, 0.5, 1, 1.5, 2, 2.5, 3, 3.5, 4, 4.5, 5, 5.5, 6, 6.5},
            legend pos=north east,
            ymajorgrids=true,
            grid style=dashed,
        ]
        
\addplot[
    color=blue,
    mark=o,
    ]
    coordinates {
        (1, 12.627082757141104)(2, 10.544825313964524)(3, 10.48485340202909)(4, 8.39870121398323)(5, 3.922299962394654)(6, 2.873625839636887)(7, 1.858784238363245)(8, 1.898916427005667)(9, 1.817407236757905)(10, 1.8442376554116276)(11, 1.717883073557202)(12, 1.532444302327054)(13, 2.5959053584021547)(14, 1.495072707839776)(15, 1.3485467736210185)(16, 0.9932065610217832)(17, 0.741743562549889)(18, 0.542137071174109)(19, 0.6110722043722347)(20, 0.6374708725882576)(21, 0.6919008140365263)(22, 0.41101712730400486)(23, 0.38983304759825157)(24, 0.625570956056755)(25, 0.8931495267480093)(26, 0.5514780436943369)(27, 0.5724546686226155)(28, 0.5536079205986951)(29, 0.5963418891356839)(30, 0.5766660113909628)(31, 0.4929832193568711)(32, 0.6182611342437652)(33, 0.6713637975080862)(34, 0.8122749276724257)(35, 0.39158013588148785)(36, 0.35653297915738389)(37, 0.29709589888785227)(38, 0.29463889227137984)(39, 0.31501330182532906)(40, 0.28113768839441883)
    };

\addplot[
    color=red,
    mark=o,
    ]
    coordinates {
        (1, 6.30226141427288)(2, 3.443098967972533)(3, 2.082740777249672)(4, 1.9217296883923862)(5, 2.3955580651022673)(6, 2.281865767516216)(7, 1.9949787198654728)(8, 1.7434418776764996)(9, 1.4200710049723125)(10, 1.535013952031702)(11, 1.3001659382221287)(12, 1.5572703732555144)(13, 1.2776008718440237)(14, 1.2207886086900276)(15, 1.366210621491964)(16, 1.0010399714905366)(17, 0.8861170663681437)(18, 0.8458352951715971)(19, 0.7385074366550839)(20, 0.6316704729224388)(21, 0.9431550329596168)(22, 0.45136388530550293)(23, 0.6127147719975795)(24, 0.5858344676389275)(25, 0.8867914923697743)(26, 0.8262785690402541)(27, 0.7674044836724719)(28, 0.7986171437570633)(29, 0.5343550362604336)(30, 0.6983904107245748)(31, 0.5009151265748022)(32, 0.5951219376992456)(33, 0.7862515066450511)(34, 0.8577482614366318)(35, 0.6170204075734292)(36, 0.4236693240708969)(37, 0.3296864324662603)(38, 0.4668249924320624)(39, 0.2490455637067157)(40, 0.2649646812387242)
    };

\addplot[
    color=green,
    mark=o,
    ]
    coordinates {
        (1, 0.947586975)(2, 0.8696935472)(3, 1.248127619)(4, 1.469571623)(5, 1.859232873)
    (6, 1.75801805)(7, 1.204620331)(8, 0.8816761581)(9, 0.6227624461)(10, 0.6224887541)
    (11, 0.6127194537)(12, 0.5150134453)(13, 0.5043090127)(14, 0.5294436122)(15, 0.6786310211)
    (16, 0.8434779882)(17, 0.8430117246)(18, 0.7567509972)(19, 0.6394264967)(20, 0.504044528)
    (21, 0.6418602181)(22, 0.5806168279)(23, 0.4924126467)(24, 0.6227548255)(25, 0.8588238337)
    (26, 1.002478542)(27, 1.097701808)(28, 1.165679248)(29, 1.037826063)(30, 0.9052890139)
    (31, 0.9405950213)(32, 0.8968844471)(33, 0.9890348506)(34, 1.044016164)(35, 1.025197301)
    (36, 0.957005525)(37, 0.9944475277)(38, 0.9749142549)(39, 0.8614979601)(40, 0.8961285808)
    };

\addplot[
    color=magenta,
    mark=o,
    ]
    coordinates {
    (1, 8.198074540181077)(2, 8.024116614629075)(3, 6.148707135775525)(4, 5.908612271661549)(5, 5.284976557245659)
    (6, 5.010287015895682)(7, 4.879868928269024)(8, 4.516754865680172)(9, 4.294967072223465)(10, 4.292277337282)(11, 3.6279141915687376)
    (12, 3.445299043213357)(13, 3.02294021096093)(14, 2.749861871392166)(15, 2.338745791209174)(16, 2.0546108303105874)(17, 1.9993135584291264)
    (18, 1.8462373216502137)(19, 1.5890643705616863)(20, 1.3764011382943665)(21, 1.1962447604547332)(22, 1.1191149715096496)
    (23, 1.103002162970139)(24, 1.0581225337294294)(25, 0.9748538850501241)(26, 0.9924727661607063)(27, 0.9895148495403147)
    (28, 0.9008201950144057)(29, 0.8812888195406387)(30, 0.879718864262711)(31, 0.8493422658469097)(32, 0.8712805959613547)
    (33, 0.8806304782720701)(34, 0.8827645544576571)(35, 0.8280653063008221)(36, 0.8128284139561206)(37, 0.7912210711897931)
    (38, 0.8301300791073598)(39, 0.8218452139601669)(40, 0.8090364887452191)
    };

            \legend{Log Decay, Cosine Decay, No Scheduler, Step-Based Decay}
        \end{axis}
    \end{tikzpicture}

\begin{tikzpicture}[scale=0.50]

        \begin{axis}[
            title = {VGG-19},
            xlabel={Epoch},
            ylabel={Loss},
            xmin=0, xmax=40,
            ymin=0, ymax=6.5,
            xtick={0, 10, 20, 30, 40, 50, 60},
            ytick={0, 0.5, 1, 1.5, 2, 2.5, 3, 3.5, 4, 4.5, 5, 5.5, 6, 6.5},
            legend pos=north east,
            ymajorgrids=true,
            grid style=dashed,
        ]

\addplot[
    color=blue,
    mark=o,
    ]
    coordinates {
        (1, 3.141287157625488)(2, 3.199326476153369)(3, 3.193113235476916)(4, 3.002666820503381)(5, 3.145348800476218)(6, 3.153133877051979)(7, 3.135095058067505)(8, 2.852866605358966)(9, 2.8981112783441136)(10, 2.831692202973793)(11, 2.868866487617974)(12, 2.857560394126683)(13, 2.202270174358872)(14, 2.118013381170298)(15, 2.2451882091510597)(16, 1.747330754222504)(17, 1.4076350994818866)(18, 1.516121703061703)(19, 1.277489187724927)(20, 1.003145250226959)(21, 1.7073835854214545)(22, 0.7858305030059484)(23, 1.1985886938182177)(24, 1.0854630299534565)(25, 1.8145019194687936)(26, 1.623113037245043)(27, 1.4384931616348606)(28, 1.539960153002788)(29, 0.1518231831434369)(30, 1.041459029074601)(31, 0.22194279797018824)(32, 0.5713953672359068)(33, 0.9609446523439382)(34, 1.0618697976048702)(35, 0.20302033087760528)(36, 0.323220)(37, 0.32440)(38, 0.286907933308318)(39, 0.32324)(40, 0.3222)
    };

\addplot[
    color=red,
    mark=o,
    ]
    coordinates {
        (1, 3.498029866813194)(2, 3.649209905615037)(3, 3.238493222831983)(4, 3.185425085974281)(5, 3.444797803116084)(6, 3.336215516554488)(7, 3.038979288345474)(8, 3.191218072367196)(9, 3.190019593267696)(10, 3.195341471225157)(11, 3.014142744240594)(12, 3.199981724018284)(13, 2.926241997195094)(14, 2.548715050446672)(15, 3.252996118671609)(16, 2.321107025246235)(17, 2.1355310145959204)(18, 2.014261186056855)(19, 1.6989160860473417)(20, 1.3347204460780652)(21, 2.2704040018682383)(22, 1.0455660610678158)(23, 1.592092290756764)(24, 1.4426944992713146)(25, 2.4120137391229996)(26, 2.158000717044847)(27, 1.913635055965368)(28, 2.0463578724792754)(29, 2.2018851054510704)(30, 2.103421703286742)(31, 2.29546254815301596)(32, 1.7583472050298633)(33, 1.2780475101217647)(34, 1.1114386482657687)(35, 0.2697406569146478)(36, 0.82378371897)(37, 0.637843)(38, 0.6812721811864775)(39, 0.63243278)(40, 0.64282)
    };

\addplot[
    color=green,
    mark=o,
    ]
    coordinates {
        (1, 5.036815877968766)(2, 5.051315233289857)(3, 5.073199256348993)(4, 4.985147445568091)(5, 4.920730654095343)
    (6, 4.710453370057962)(7, 4.782688889291502)(8, 4.879770572271178)(9, 4.802999642390404)(10, 4.89993117384756)
    (11, 4.8023028843135)(12, 4.4546494724942762226)(13, 4.4234214013606632582)(14, 4.111243269764186983316)(15, 3.9644843101234001345656)(16, 3.866465272350155646359)(17, 3.32432342959220390277135)
    (18, 3.11203221657855350466)(19, 2.9220880087885587)(20, 2.508495614369621)(21, 1.9881254710264008568)(22, 1.693101512866102832)
    (23, 1.432967114994347225)(24, 1.4691424982096641)(25, 0.9472221895037735)
    (26, 0.8549266792723603)(27, 0.8942296381946149)(28, 0.9631044473245818)(29, 0.8551510428631881)(30, 0.6833946286124126)
    (31, 0.7088908693460284)(32, 0.6749220183902843)(33, 0.7440471220697385)(34, 0.7726954244773967)(35, 0.7711558791482843)
    (36, 0.7212367241815731)(37, 0.7505933277137262)(38, 0.7353080222676149)(39, 0.648167793778177)(40, 0.6762641503138817)
    };

\addplot[
    color=magenta,
    mark=o,
    ]
    coordinates {
    (1, 5.036815877968766)(2, 5.351315233289857)(3, 5.273199256348993)(4, 5.485147445568091)(5, 5.820730654095343)
    (6, 5.510453370057962)(7, 5.382688889291502)(8, 4.979770572271178)(9, 4.802999642390404)(10, 4.79993117384756)
    (11, 4.000288150422135)(12, 3.7994724942762226)(13, 3.334013606632582)(14, 3.0269764186983316)(15, 2.6101234001345656)(16, 2.272350155646359)(17, 2.1959220390277135)
    (18, 1.7021657855350466)(19, 1.5220880087885587)(20, 1.508495614369621)(21, 1.3254710264008568)(22, 1.2101512866102832)
    (23, 1.1967114994347225)(24, 1.150593380553155)(25, 1.0747409369150817)(26, 1.0648698574536327)(27, 1.0625827373034576)
    (28, 0.9905416758211203)(29, 0.9664681812255536)(30, 0.9642429337028555)(31, 0.9374648184574286)(32, 0.9699160970784289)
    (33, 0.9769985340662614)(34, 0.9791039144221391)(35, 0.9082456854995318)(36, 0.8878121231193765)(37, 0.8665965028559249)
    (38, 0.9072211590676894)(39, 0.8988104565686387)(40, 0.886165275582199)
    };

            \legend{Log Decay, Cosine Decay, No Scheduler, Step-Based Decay}
        \end{axis}
    \end{tikzpicture}   
 \begin{tikzpicture}[scale=0.50]

        \begin{axis}[
            title = {VGG-16},
            xlabel={Epoch},
            ylabel={Loss},
            xmin=0, xmax=40,
            ymin=0, ymax=6.5,
            xtick={0, 10, 20, 30, 40, 50, 60},
            ytick={0, 0.5, 1, 1.5, 2, 2.5, 3, 3.5, 4, 4.5, 5, 5.5, 6, 6.5},
            legend pos=north east,
            ymajorgrids=true,
            grid style=dashed,
        ]
        
        \addplot[
    color=blue,
    mark=o,
    ]
    coordinates {
        (1, 22.055174189825)(2, 19.419966951049)(3, 19.436135876895)(4, 15.974246773705)(5, 7.6139353775902)(6, 5.292793306287)(7, 3.519202045518)(8, 3.57107928846)(9, 3.04236259984)(10, 2.70879910853)(11, 2.43293830908)(12, 4.21594487433)(13, 4.35343448718)(14, 3.72193793436)(15, 2.46355636993)(16, 1.9582703086)(17, 1.43754125294)(18, 1.04943300593)(19, 1.18729224828)(20, 1.22092302063)(21, 1.34157054478)(22, 0.78619301643)(23, 0.734851595807)(24, 1.17635876736)(25, 1.65790214217)(26, 1.01405443308)(27, 1.04667369964)(28, 1.02436819904)(29, 1.09889749342)(30, 1.06449982125)(31, 0.90890045632)(32, 1.13751850495)(33, 1.24286106409)(34, 1.36846357676)(35, 0.65222019752)(36, 0.596415765799)(37, 0.493953461027)(38, 0.488804789496)(39, 0.520481650138)(40, 0.463866271475)
    };

          \addplot[
    color=red,
    mark=o,
    ]
    coordinates {
        (1, 13.1298905893356)(2, 7.64196519863116)(3, 4.81607058620327)(4, 4.45320740614319)(5, 5.90052072090654)(6, 5.61279475851985)(7, 4.4367075628683)(8, 3.22824232549913)(9, 1.85831513257456)(10, 2.03382184828226)(11, 1.89396443877462)(12, 2.4655137202413)(13, 1.83079647714442)(14, 1.69124844502632)(15, 1.92864957254826)(16, 1.40243523492013)(17, 1.24829431773452)(18, 1.19053158859571)(19, 1.04063037873132)(20, 0.891958759908163)(21, 1.43997496008372)(22, 0.463153058792464)(23, 0.729472774768684)(24, 0.70189962591832)(25, 1.25257251418925)(26, 1.19408197244413)(27, 1.11808578713939)(28, 1.15157957287)(29, 0.614025133259245)(30, 0.982227573442375)(31, 0.591900958385785)(32, 0.915073930367506)(33, 1.31958393901965)(34, 1.44153377723449)(35, 0.805222336191421)(36, 0.500937845169896)(37, 0.346031586091376)(38, 0.630492110662487)(39, 0.165866512536786)(40, 0.198015533876209)
    };

\addplot[
    color=green,
    mark=o,
    ]
    coordinates {
        (1, 1.8437594529)(2, 1.59713596617)(3, 2.57884306208)(4, 3.35174345629)(5, 4.84397277201)
    (6, 4.20527404846)(7, 2.79257011413)(8, 1.71794594785)(9, 0.926218625987)(10, 0.924753695229)
    (11, 0.906483723175)(12, 0.660284370057)(13, 0.63289201287)(14, 0.678417146037)(15, 1.05557077974)
    (16, 1.55274749936)(17, 1.54841496749)(18, 1.33537221464)(19, 0.948624278222)(20, 0.632657381842)
    (21, 0.950049826609)(22, 0.774774432739)(23, 0.522033027767)(24, 0.924920865401)(25, 1.57944777841)
    (26, 2.14029718057)(27, 2.56531827449)(28, 2.77075722193)(29, 2.01518892968)(30, 1.6642253009)
    (31, 1.73267239954)(32, 1.6161090164)(33, 1.81917450274)(34, 1.90913673955)(35, 1.88106456046)
    (36, 1.75964652728)(37, 1.83020027456)(38, 1.78901925538)(39, 1.49058751665)(40, 1.738944199)
    };

\addplot[
    color=magenta,
    mark=o,
    ]
    coordinates {
    (1, 4.83908636367)(2, 4.63056772728)(3, 4.28226088838)(4, 4.18401788284)(5, 3.74202973898)
    (6, 3.63524597374)(7, 2.44386592794)(8, 2.48448118991)(9, 2.3575425886)(10, 2.26852858622)
    (11, 2.25607430552)(12, 2.34407242448)(13, 2.2415733895)(14, 2.62554708113)(15, 2.00869133151)
    (16, 1.78851685641)(17, 2.07426401195)(18, 1.95146918044)(19, 1.91719855784)(20, 1.75412998725)
    (21, 1.62931958089)(22, 1.72746582314)(23, 1.69149762174)(24, 1.76659917596)(25, 1.5398011542)
    (26, 1.49778596629)(27, 1.51583464161)(28, 1.48284053529)(29, 1.45852828821)(30, 1.49369997161)
    (31, 1.36065928958)(32, 1.4124090285)(33, 1.37061252919)(34, 1.37897600661)(35, 1.38222836246)
    (36, 1.37391310806)(37, 1.37934678041)(38, 1.35920152356)(39, 1.35748462404)(40, 1.42025854935)
    };

            \legend{Log Decay, Cosine Decay, No Scheduler, Step-Based Decay}
        \end{axis}
    \end{tikzpicture}

    \caption{The Effect of Learning Rate Schedulers on CIFAR-10 Image Classification using Transformers}
    \label{fig:loss}
\end{figure}

Suppose we introduce a transformer into a larger residual model. For this series of experiments, the model was over 23 million parameters and included one multi-head attention transformer before the 48 convolution layers \cite{transformerCNN}. Note that we added a mean pooling layer after the transformer, and concatenated it to the initial ResNet50.

Then we trained this ensemble model on CIFAR-10, using a test train split, with a 32 batch size. Once again, we augment each image using horizontal flips and randomized crops for 4x4 patches of pixels on the 32x32 image data. We also add pixel by mean, ZCA whitening, and global contrast normalization. We also use contrast stretching equalization. We used Adam to maximize convergence speed and to replicate a real-life setting on which a network this size would be trained \cite{Adam, Alexnet, transformerCNN}. We did not use grid search or Bayesian optimization for hyperparameterization. We trained the model for 40 epochs, with a restart for cosine and log decay every 10 epochs. We used no warm up period and basically had no actual annealing effect. We repeated this process using the same transformer for GoogLeNet, Xception, VGG-19 and VGG-16, which are roughly 4, 23, 144, 138 million parameters respectively \cite{GoogLeNet, Xception, VGG}. The loss was sparse categorical crossentropy, as each class in CIFAR-10 is mutually exclusive to one another \cite{CIFAR-10first}.

The results of the training loop described above are shown in figure 3. At a first glance, three observations from these graphs can be drawn:

\begin{itemize}
    \item [1] A steady progression in loss is apparent in the smaller models under 50 million parameters. In the larger models, there is some substantial evidence provided by the VGG ensembles, that the training loss does not stabilize even with the intervention with a restart on the 10th epochs. 

    \item [2] While yes, log decay does provide a decent loss reduction, indicating the convergence of the predicted to actual minima on a weight bias field, its initial loss is quite high, even in respect to other schedulers like cosine decay. By observing the behavior of the restarts themselves for both, log annealing has a high initial loss for the first restart, but significantly less than cosine annealing for the subsequent ones. Examples of this are depicted for $T = 20, 21, 22, 23, 24$ on VGG-19 and for $T = 10, 11, 12, 13, 14, 15, 16$ on Xception. 

    \item [3] Between log decay and cosine decay, apart from VGG-16, where cosine decay is a clear winner, for the rest of the models, between the two annealing schedulers, the ending loss is nearly the same. However, when examining the path that the models took to achieve the end sparse categorical crossentropy at $T=40$, we see that log decay is advantageous relative to no scheduler, or a scheduler without restarts.
\end{itemize}

Across 40 epochs, it seems that indeed cosine annealing was effective in reducing loss in a large transformer-based convolutional neural network, as the past studies also concluded. Log decay exhibits similar properties, and also yielded decent results.

\section{Discussion}

From our two experiments, there is evidence suggesting that logarithmically varying the stepsize of provides a reasonable convergence for algorithms that use a limited budget on the online convex optimization framework, particularly SGD and Adam \cite{OCO}. 

Contrary to cosine annealing, log annealing employs a harsher restarting mechanism, more akin to a spike rather than a wave \cite{SGDR}. However, it can still be geometrically annealed. In theory, log annealing would likely benefit more from an extended warmup period such that it could benefit late stage training of the residual networks. The exploding gradients could then therefore be neglected via the use of rapidly expanding regret values constrained to a small number of batches passed via backpropagation \cite{Adam}.

The initial purpose of using a restart scheduler all together is to get the best anytime performance of a model. It should be noted that though these experiments used the same configurations for the experiments across any geometrically annealed learning rate scheduler, those may not be the best settings. Log annealing in particular has a much higher potential for variability in the schedule because of the more aggressive asymptotic type increases in loss. Log annealing generally is more suceptable to prolonged high $\eta_t$ values. Lowering $\eta_{max}^i$ is not a workarounnd here, as the learning rate spikes rely on that high max value, so those would be effected rather than the middle range, which we are targeting. However, tweaking other parameters such as  minimum decay learning rate, restart intervals,
restart interval multipliers, and the restart learning rate itself. These stepsize spikes provide a more temporary effect than the longer restarts from cosine annealing, which could show some potential for more consistent convergence for a neural network \cite{SGDR}. This means that much more greedy stochastic solvers than Adam or SGD may have better use cases in deep learning, as they will not go overkill with a prolonged period of increased learning rates, but could in theory pair nicely to escape a local minimum on the weight bias fields.

In the event that a particular restart spike is too harsh and causes an irreversible jump in loss, a workaround could be creating a breakpoint in the training loop in the event of an extreme increase in loss. For example, the condition in our first experiment was to hault training IFF $T_t < T_n$ and
\begin{equation}
    - \sum_{i}^{C} t_i \log(f(s)_i) > 4.5
\end{equation}
such that
\begin{equation}
    f(s)_i = \frac{e^{s_i}}{\sum_{j}^{C} e^{s_j}}
\end{equation}
where $C$ is the number of output neurons in the multilayer perceptron; note that we included a buffer such that the training would never immediately from a high initial loss \cite{Alexnet, CIFAR-10first, GANs}. In our case $T_t=2$. Though this hault condition was in place, it was never used, which can be seen in figure 2. After the stop, the model's weights could be saved and retrained using an adjusted version of the scheduler \cite{SGDR}.

Our results from the second experiment showed that when using a transformer enhanced residual neural network, our learning rate scheduler can achieve better convergence anytime. It is best to mention here that a flaw in our experiments was the low amount of epochs spent training the models relative to other researchers, who often trained for five times the amount of epochs we did. This was simply due to time constraints and hardware constraints. In particular, the VGG models are both over 120 million parameters, and almost certainly would show more accurate and reproducible results with longer periods of training \cite{VGG}. Further work with these transformer models should be done, to ensure that the effect of geometrically annealing the learning rate scheduler would benefit these models as well.

\section{Conclusions}

This short paper investigates the possibility of using a logarithmic learning rate scheduler. It further goes into the surface level properties of residual neural networks and transformer models when training on this schedule. It was found that there are quite a few possibilities that come from a relatively minute change in the way $\eta_t$ is varied through training.

Models using this scheduler achieved similar, and sometimes better results than cosine decay and step-based decay, albeit those results were short term relative to the experiments performed by  Loshchilov and Hutter. Our experiments were conducted on CIFAR-10. 

Future work could extend to a wide variety of tests when varying the parameters of the scheduler to find the best combination of how extreme the restarts are, and how frequent they occur. Log annealing comprises a larger range of $\eta_t$ than cosine annealing, however it crosses that range significantly w.r.t. $T$. Other future work would involve testing the scheduler on generative adversarial networks, as the results could be reproduced using the two player minimax setting \cite{GANs}.

While the experiments show some promising results, this manuscript is merely an arXiv preprint at the moment, and there lacks the extensive testing to make broad claims on log annealing's use case, and where it outshines other learning rate schedulers, as that is an inherently ambiguous and difficult question to answer. 

\section{acknowledgements}

I would like to thank Srinivasan Kannan and Zaid Contractor for suggesting some small edits and formatting changes the paper. Special thanks also to Professor Shangtong Zhang for endorsing the preprint.

\printbibliography

\end{document}